\documentclass[letterpaper, 10 pt, journal, twoside]{IEEEtran}

\IEEEoverridecommandlockouts                              

\usepackage{times}
\usepackage{epsfig}
\usepackage{graphicx}
\usepackage{amssymb} 
\usepackage{amsmath}
\usepackage[linesnumbered,ruled]{algorithm2e}
\usepackage{array}
\usepackage{rotating}
\usepackage{multirow}
\usepackage{csquotes}
\usepackage{color}
\usepackage{booktabs}
\usepackage{cite}
\usepackage{amsmath,amssymb,amsfonts}
\usepackage{algorithmic}
\usepackage{graphicx}
\usepackage{url}
\usepackage{textcomp}
\usepackage[dvipsnames]{xcolor}
\usepackage{hyperref}

\newcolumntype{P}[1]{>{\centering\arraybackslash}p{#1}}

\newcommand{\thickhline}{%
    \noalign {\ifnum 0=`}\fi \hrule height 1pt
    \futurelet \reserved@a \@xhline
}

\begin{document}

\title{
3D-MiniNet: Learning a 2D Representation from Point Clouds  for Fast and Efficient 3D LIDAR Semantic Segmentation}

\author{I\~{n}igo~Alonso$^{1}$, Luis Riazuelo$^{1}$,  Luis Montesano$^{1,2}$, and  Ana C.~Murillo$^{1}$%
\thanks{This paper was recommended for publication by Tamim Asfour upon
evaluation of the Associate Editor and Reviewers' comments. 
This project was partially funded by projects FEDER/ Ministerio de Ciencia, Innovaci{\'o}n y Universidades/ Agencia Estatal de Investigaci{\'o}n/RTC-2017-6421-7, PGC2018-098817-A-I00 and PID2019-105390RB-I00, Arag{\'o}n regional government (DGA T45 17R/FSE) and the Office of Naval Research Global project ONRG-NICOP-N62909-19-1-2027.}
\thanks{$^{1}$ RoPeRt group, at DIIS - I3A, Universidad de Zaragoza, Spain. {\tt\small  \{inigo, riazuelo, montesano, acm\}@unizar.es}}
\thanks{$^{2}$ Bitbrain, Zaragoza, Spain} 
}

\maketitle

\begin{abstract}

LIDAR semantic segmentation is an essential task that provides 3D semantic information about the  environment to robots.
Fast and efficient semantic segmentation methods  are needed to match the strong computational and temporal restrictions of many real-world robotic  applications. 
This work presents 3D-MiniNet, a novel approach for LIDAR semantic segmentation that combines 3D and 2D learning layers. 
It first learns a 2D representation from the raw points through a novel projection which extracts local and global information from the 3D data. 
This representation is fed to an efficient 2D Fully Convolutional Neural Network (FCNN) that produces a 2D semantic segmentation. These 2D semantic labels are re-projected back to the 3D space and enhanced through a post-processing module. 
The main novelty in our strategy relies on the projection learning module. Our detailed ablation study shows how each component contributes to the final performance of 3D-MiniNet. 
We validate our approach on well known public benchmarks (SemanticKITTI and KITTI), where 3D-MiniNet gets state-of-the-art results while being faster and more parameter-efficient than previous methods. 
\end{abstract}


\section{INTRODUCTION}
\IEEEPARstart{A}{utonomous}  robotic systems use sensors to perceive the world around them. RGB cameras and LIDAR are very common due to the essential data they provide. 
 One of the key building blocks of autonomous robots is semantic segmentation. Semantic segmentation assigns a class label to each LIDAR point or camera pixel.  This detailed semantic information is essential for decision making in real-world dynamic scenarios. 
 LIDAR semantic segmentation provides very useful information to autonomous robots when performing tasks such as Simultaneous Localization And Mapping (SLAM) \cite{jian2019semantic, zhao2019lidar}, autonomous driving \cite{milioto2019rangenet++} or inventory tasks \cite{chen2020sloam}, especially for identifying dynamic objects.
 In these scenarios, it is critical to have models that provide accurate semantic information in a  fast and efficient manner, which is particularly challenging working with 3D LIDAR data. 
 On one hand, the commonly called \textit{point-based approaches} \cite{qi2017pointnet++,qi2017pointnet,tangentconv} tackle this problem directly executing 3D point-based operations, which is computationally expensive to operate at high frame rates. On the other hand, approaches that project the 3D information into a 2D image (\textit{projection-based approaches}) are more efficient  \cite{wu2019squeezesegv2,milioto2019rangenet++,wu2018squeezeseg, wang2018pointseg, dewan2019deeptemporalseg} but do not exploit the raw 3D information. 
 Recent results on fast \cite{milioto2019rangenet++} and parameter-efficient \cite{zhang2019shellnet} semantic segmentation models are facilitating the adoption of semantic segmentation in real-world robotic applications \cite{li2020integrate, behley2019semantickitti}.

\begin{figure}[!t]
\centering
\includegraphics[width=1\linewidth]{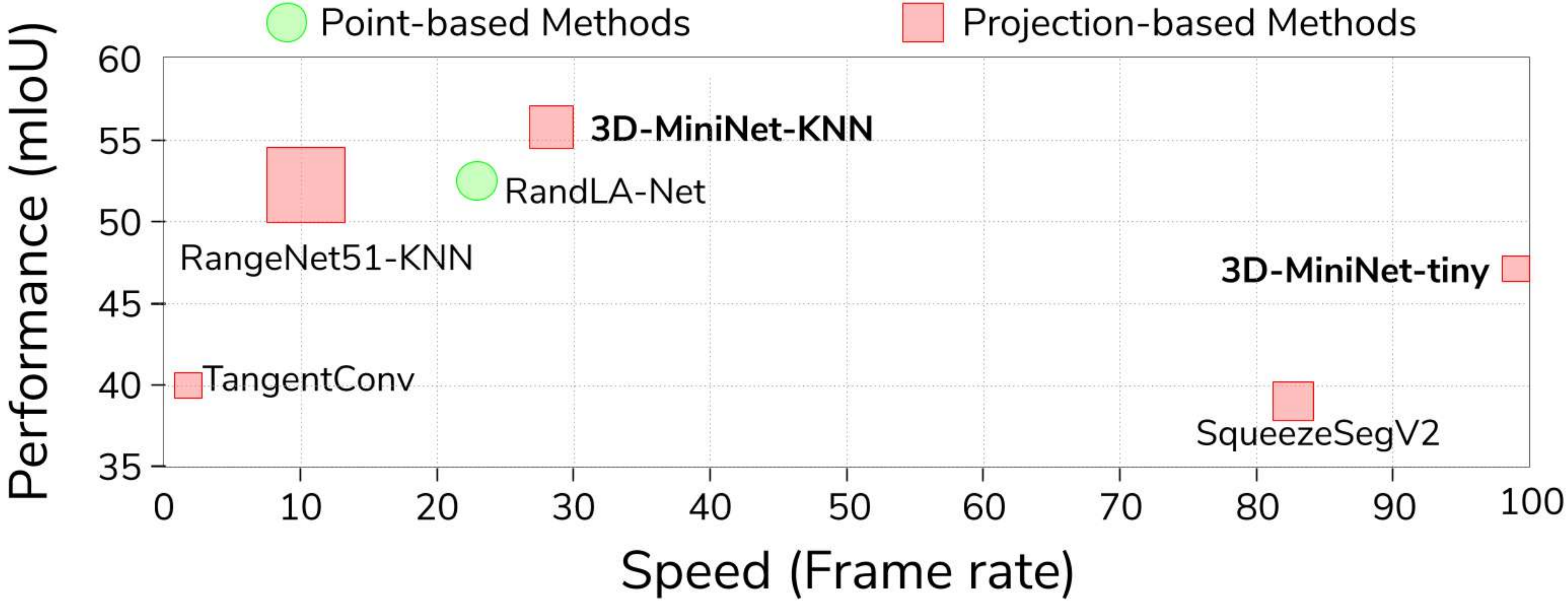}
\caption{3D LIDAR semantic segmentation \textbf{accuracy vs speed} on SemanticKITTI test set \cite{behley2019semantickitti}.
\textcolor{Green}{Green} circles depict point-based methods and \textcolor{Red}{red} squares are projection-based methods. Area of these shapes is proportional to the method number of parameters. The proposed 3D-MiniNet outperforms previous methods with less parameters and faster execution. 
}
\label{fig:fast-efficient}
\end{figure} 

This work presents a novel fast and parameter-efficient approach for 3D LIDAR semantic segmentation that consists of three modules (as detailed in Sec. \ref{sec:method}). The main contribution relies on our 
3D-MiniNet module. 
3D-MiniNet runs the following two steps: (1) It learns a 2D representation from the 3D point cloud (following previous works on 3D object detection \cite{lang2019pointpillars, zhou2018voxelnet, zhou2020end}); (2) It computes the segmentation through a fast 2D fully convolutional neural network.

Our best configuration achieves state-of-the-art results in well known public benchmarks (SemanticKITTI \cite{behley2019semantickitti} and KITTI dataset \cite{geiger2012wekitti}) while being faster and more parameter efficient that prior work. Figure \ref{fig:fast-efficient} shows how 3D-MiniNet achieves better precision-speed trade-off than previous methods. 
The main novelties with respect to existing approaches, that facilitate these improvements, are:
\begin{itemize}
\item An extension of MiniNet-v2 for 3D LIDAR semantic segmentation: 3D-MiniNet.
\item Our novel projection module.
 
\item A validation of 3D-MiniNet on the SemanticKITTI benchmark \cite{behley2019semantickitti} and KITTI dataset \cite{geiger2012wekitti}. 

\end{itemize}


The proposed projection module learns a rich 2D representation through different operations. It consists of four submodules: a context feature extractor, a local feature extractor, a spatial feature extractor and the feature fusion. We provide a detailed ablation study on this module showing how each proposed components contributes to improve the final performance of 3D-MiniNet.
Besides, we implemented a fast version of the point neighbor search based on a sliding-window on the spherical projection \cite{lunet} in order to compute it at an acceptable frame-rate. All the code and trained models are available online~\footnote{\url{https://sites.google.com/a/unizar.es/semanticseg/}}.

\section{RELATED WORK}

\subsection{2D Semantic Segmentation}

Current 2D semantic segmentation state-of-the-art methods are deep learning solutions \cite{deeplabv3plus2018,chen2017rethinking,jegou2017one,long2015fully}. Semantic segmentation architectures are evolved from convolutional neural networks (CNNs) architectures for classification tasks, adding a decoder on top of the CNN.
Fully Convolutional Neural Networks for Semantic Segmentation (FCNN) \cite{long2015fully} carved the path for modern semantic segmentation architectures. The authors of this work propose to upsample the learned features of classification CNNs using bilinear interpolation up to the input resolution and compute the cross-entropy loss per pixel. 
Another of the early approaches, 
SegNet \cite{badrinarayanan2017segnet}, proposes a symmetric encoder-decoder structure using the unpooling operation as upsampling layer. 
More recent works improve these earlier segmentation architectures by adding novel operations or modules proposed initially within CNNs architectures for classification tasks. FC-DenseNet \cite{jegou2017one} follows DenseNet work \cite{huang2017densely} using dense modules. PSPNet \cite{zhao2017pyramid} uses ResNet \cite{he2016deep} as its encoder and introduces the Pyramid Pooling Module incorporated at the end of the CNN allowing to learn effective global contextual priors. Deeplab-v3+ \cite{deeplabv3plus2018} is one of the top-performing architectures for segmentation.
Its encoder is based on Xception \cite{chollet2016xception}, which makes use of depthwise separable convolutions \cite{sifre2014rigid} and  atrous (dilated) convolutions \cite{YuKoltun2016}.

With respect to efficiency, 
ENet \cite{paszke2016enet} set up certain basis which following works, such as ERFNet \cite{romera2018erfnet}, ICNet \cite{zhao2018icnet}, have built upon. The main idea is to work at low resolutions, i.e., quick downsampling, and to focus the computation on the encoder having a very light decoder. 
MiniNetV2 \cite{alonso2020MininetV2} uses a multi-dilation depthwise separable convolution, which efficiently learns both local and global spatial relationships. 
In this work, we take MiniNetV2 as our backbone and adapt it to capture information from raw LIDAR points.

\subsection{3D Semantic Segmentation}
There are three main groups of strategies to approach this problem: point-based methods, 3D representations and projection-based methods. 

\subsubsection{Point-based Methods}
Point-based methods work directly on raw point clouds. The order-less structure of the point clouds prevents standard CNNs to work on this data. 
The pioneer approach and base of the following point-based works is PointNet \cite{qi2017pointnet}. PointNet proposes to learn per-point features through shared MLP (multi-layer perceptron) followed by symmetrical pooling functions to be able to work on unordered data.
Lots of works have been later proposed based on PointNet.
Following with the point-wise MLP idea, PoinNet++ \cite{qi2017pointnet++} groups points in an hierarchical manner and learns from larger local regions. The authors also propose a multi-scale grouping for coping with the non-uniformity nature of the data.
In contrast, other approaches propose different types of operations following the convolution idea. Hua et al.\cite{hua2018pointwise} propose to bin neighboring points into kernel cells for being able to perform point-wise convolutions.
Other works resort to graph networks to capture the underlying geometric structure of the point cloud. Loic et al. \cite{landrieu2018large} use a directed graph to capture the structure and context information. For this, the authors represent the point cloud as a set of interconnected superpoints.

\subsubsection{3D representations}
There are different kinds of representations of the raw point cloud data which have been used for 3D semantic segmentation. SegCloud \cite{tchapmi2017segcloud} makes use of a \textit{volumetric or voxel representation}, which is a very common way for encoding and discretizing the 3D space.  This approach feeds the 3D voxels into a 3D-FCNN \cite{long2015fully}. Then, the authors introduce a deterministic trilinear interpolation to map the coarse voxel predictions back to the original point cloud and apply a CRF as a final step. The main drawback of this voxel representation is that 3D-FCNN has very slow execution times for real-time applications.
Su et al. \cite{su2018splatnet} proposed SPLATNet, making use of another type of representation: \textit{Permutohedral Lattice representation}. This approach interpolates the 3D point cloud to a permutohedral sparse lattice and then bilateral convolutional layers are applied to convolve on occupied parts of the representation. LatticeNet \cite{rosu2019latticenet} was later proposed improving SPLATNet proposing its DeformSlice module for re-projecting the lattice feature back to the point cloud.

\subsubsection{Projection-based Methods}
 This type of approaches rely on projections of the 3D data into a 2D space.
For example, TangentConv\cite{tangentconv} proposes to project the neighboring points into a common tangent plane where they perform convolutions.
Another type of projection-based method is the \textit{spherical representation}. This strategy consists of projecting the 3D points into a spherical projection and has been widely used for LIDAR semantic segmentation. This representation is a 2D projection that allows the application of 2D images operations, which are very fast and work very well on recognition tasks. SqueezeSeg \cite{wu2018squeezeseg} and its posterior improvement SqueezeSegV2 \cite{wu2019squeezesegv2}, based on SqueezeNet architecture \cite{iandola2016squeezenet}, show that very efficient semantic segmentation can be done through this projection. The more recent work from Milioto et al. \cite{milioto2019rangenet++} combines the DarkNet architecture \cite{redmon2018yolov3} with a GPU based post-processing method 
for real-time semantic segmentation. 

Projection-based approaches tend to be faster than other representations, but they lose the potential of learning 3D features.
LuNet \cite{biasutti2019lu} is a recent work which proposes to learn local features using point-based operations before projecting into the 2D space.  
 Our novel projection module tackles with this issue by including a context feature extractor based on point-based operations. Besides, we build a faster and more parameter-efficient architecture and a faster implementation of LuNet's neighbor search method.

\begin{figure}[!tb]
\centering
\includegraphics[width=1\linewidth]{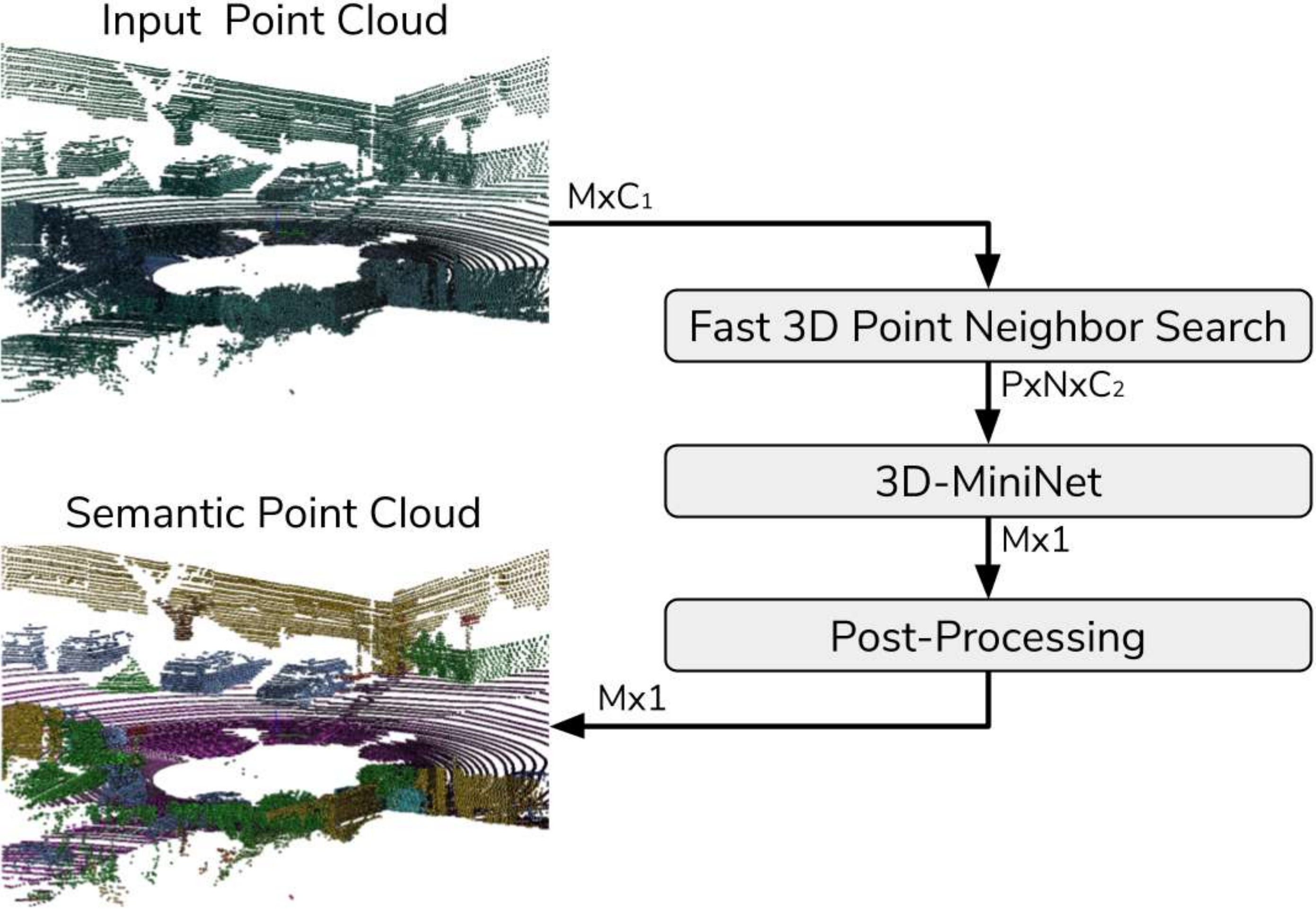}
\caption{Proposed approach overview. 
The $M$ points from the input point cloud (with $C_\mathrm{1}$  features) are split into $P$ groups of $N$ points with our fast 3D point neighbor search. Each point has a $C_{1}$ feature vector, which is extended to $C_{2}$ in this process with data relative to each group.
The proposed 3D-MiniNet takes the point groups and predicts one semantic label per point. 
A post-processing method  \cite{milioto2019rangenet++} is used to refine the final results.}
\label{fig:pipeline}
\end{figure}

\section{3d-mininet: lidar point cloud segmentation}
\label{sec:method}

Our novel approach for LIDAR semantic segmentation is summarized in Fig. \ref{fig:pipeline}. It consists of three modules: (A) fast 3D point neighbor search, (B) 3D-MiniNet, which takes $P$ groups of $N$ points and outputs the segmented point cloud and, (C) the KNN-based post-processing which refines the final segmentation.

There are two main issues that typically prevent point-based models to run at an acceptable frame-rate compared to projection-based methods: 3D point neighbor search is a required, but slow, operation and performing 3D operations is slower than using 2D convolutions. In order to alleviate these two issues, our approach includes a fast point neighbor search proxy (subsection \ref{sec:neighbor_search}), and a module to minimize expensive point-based operations, which takes raw 3D points and outputs a 2D representation to be processed with a 2D CNN (subsection \ref{sec:projection_module}).

\subsection{Fast 3D Point Neighbor Search}
\label{sec:neighbor_search}

We need to find the 3D neighbors because we want to learn features that encode the relationship of each point with their neighbors in order to learn information about the shape of the point-cloud.  
In order to perform the 3D neighbor search more efficiently, we first project 
the point cloud into a spherical projection of shape $W\times H$, mapping every 3D point ($x, y, z$) into a 2D coordinate ($u, v$), i.e., $\mathbb{R}^3 \xrightarrow{} \mathbb{R}^2$:
\begin{equation}
\label{eq:projection}
\left(\begin{array}{c}
{u} \\
{v}
\end{array}\right)=\left(\begin{array}{c}
{\frac{1}{2}\left[1-\arctan (y, x) \pi^{-1}\right]  W} \\
{\left[1-\left(\arcsin \left(z r^{-1}\right)+\mathrm{f}_{\mathrm{up}}\right) \mathrm{f}^{-1}\right] H}
\end{array}\right),
\end{equation}

\noindent where $f=f_\mathrm{up}+f_\mathrm{down}$  is the vertical field-of-view of the sensor and $r$ is the depth of each point. We perform the projection of Eq. \ref{eq:projection}  following \cite{milioto2019rangenet++}, where each pixel encodes one 3D point with five features: $C_1 = \{x, y, z, depth, remission\}$.

We perform the point neighbor search in the spherical projection space using a sliding-window approach. Similarly to a convolutional layer, we get groups of pixels, i.e., projected points, by sliding a $k\times k$ window across the image. The generated groups of points have no intersection, i.e., each point belongs only to one group. 
This step generates $P$ point groups of $N$ points each ($N=k^2$), where all points from the spherical projection are used ($P\times N=W\times H$). 

Before feeding the actual segmentation module, 3D-MiniNet, with these point groups, the features of each point are augmented.  For each group we compute the relative ($r$) feature values for each point. They are computed with respect to the group mean for each $C_1$ feature (similar to previous works which compute features relative to a center point \cite{lunet, zhang2019shellnet}). Besides, similar to \cite{zhang2019rotation}, we compute the 3D euclidean distance of each point to the mean point. 
Therefore, each point has now eleven features: $C_2 = \{ x, x_\mathrm{r}, y, y_\mathrm{r}, z, z_\mathrm{r}, depth, depth_\mathrm{r}, remission, remission_\mathrm{r}$, $ d_{Euc}\}$.

\begin{figure*}[!tb]
\centering
\includegraphics[width=1\linewidth]{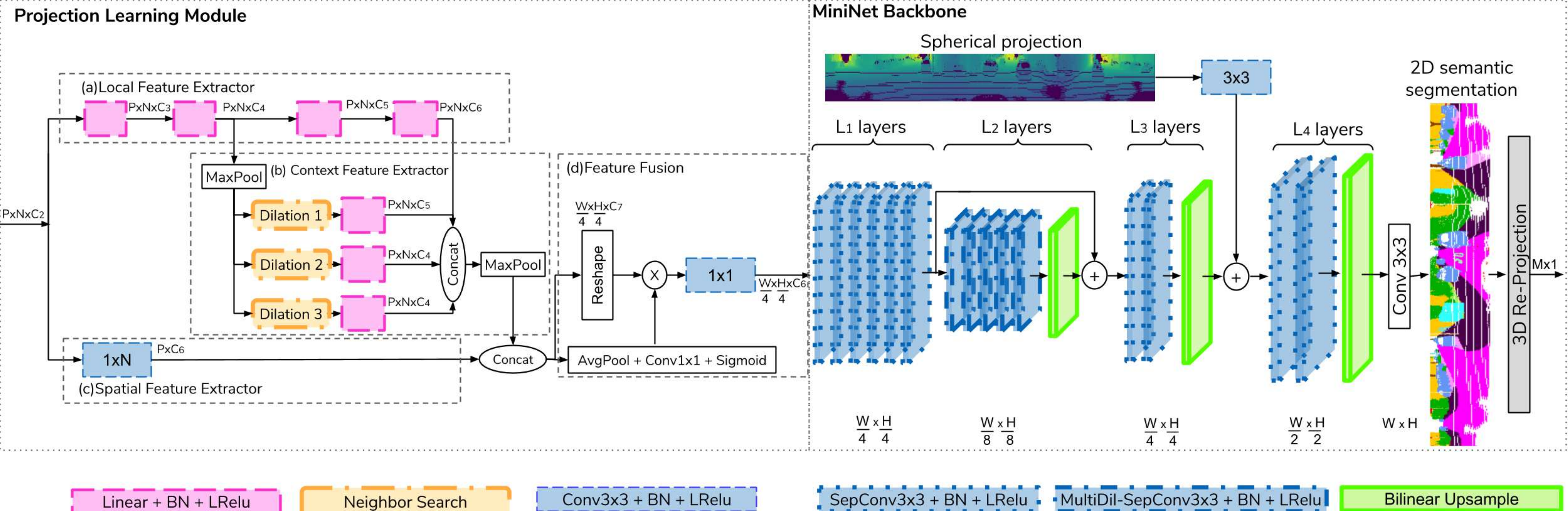}
\caption{3D-MiniNet overview. It takes $P$ groups of $N$ points each and computes semantic segmentation of the $M$ points of the point cloud where $P\times N = M$.
It consists of two main modules: our proposed learning module (on the left) which learns a 2D tensor which is fed to the second module, an efficient FCNN backbone (on the right) which computes the 2D semantic segmentation. Each 3D point of the point cloud is given a semantic label based on the 2D segmentation. Best viewed in color.}
\label{fig:mininet}
\end{figure*}

 \subsection{3D-MiniNet}
  \label{sec:3d-mininet}
 
 3D-MiniNet consists of two modules, as represented in Fig. \ref{fig:mininet}: the proposed projection module, which takes the raw point cloud and computes a 2D representation, and our efficient backbone network based on MiniNetV2 \cite{alonso2020MininetV2} to compute the semantic segmentation.
 
 \subsubsection{Projection Learning Module}
 \label{sec:projection_module}
 
The goal of this module is to transform raw 3D points to a 2D representation that can be used for efficient segmentation. The input of this module if the output of the point neighbor search described in the previous subsection. It is a set of $P$ groups, where each group contains $N$ points with $C_2$ features each, gathered through the sliding-window search on the spherical projection as explained in the previous subsection. 

The following three kinds of features are extracted from the input data 
(see left part of Fig. \ref{fig:mininet} for a visual description of this proposed module) and fused in the final module step:

\paragraph*{Local Feature Extractor}
The first feature is a PointNet-like local feature extraction (see projection learning module (a) of Fig. \ref{fig:mininet}). It runs four linear layers shared across the groups followed by a BatchNorm \cite{ioffe2015batch} and LeakyRelu \cite{maas2013rectifier}. We follow PointPillars \cite{lang2019pointpillars} implementation of these shared linear layers using $1\times1$ convolutions across the tensor resulting in very efficient computation when handling lots of point groups.

\paragraph*{Context Feature Extractor}
The second feature extraction (projection learning module (b) of Fig. \ref{fig:mininet}) learns context information from the points.This is a very important module because although  context information can be learned through the posterior CNN, point-based operations learn different features than convolutions.  Therefore, this module helps learning a richer representation with information than might not be learned through the CNN.

The input of this context feature extractor is the output of the second linear layer of the local feature extractor (giving the last linear layer as input would drop significantly the frame-rate due to the high number of features). This tensor is maxpooled (in order to complete the PointNet-like operation which work on unordered points) and then, our fast neighbor search is run to get point groups. In this case, three different groupings (using our point neighbor search) are performed with a $3\times3$ sliding window with different dilation rates of  1, 2, 3 respectively. Dilation rates, as in convolutional kernels \cite{YuKoltun2016}, keep the number of grouped points low while increasing the receptive field allowing a faster context learning. We use zero-padding and a stride of $1$ for keeping the same size. 
After every grouping we perform a linear, BatchNorm and LeakyRelu.  
The outputs of these two feature extractor modules are concatenated and applied a maxpool operation over the $N$ dimension.  This maxpool operation keeps the feature with higher response along the neighbor dimension, being order-invariant with respect to the neighbor dimension. The maxpool operation also makes the learning robust to pixels with no point information (spherical projection coordinates with no point projected).

\paragraph*{Spatial Feature Extractor}
The last feature extraction operation is a convolutional layer of kernel $1\times N$ (projection learning module (c) of Fig. \ref{fig:mininet}).   Convolutions can extract features of each point with respect to the neighbors when there is an underlying spatial structure which is the case,  as the point groups are extracted from a 2D spherical projection.
In the experiment section, we take this feature extractor as our baseline without the two others which is equivalent of performing only standard convolutions on the spherical projection.

\paragraph*{Feature Fusion}
Lastly, a feature fusion with self-attention module is applied. 
It learns to reduce the feature space into an specified number of features, learning which features are more important.
It consists of three stages: (1) concatenation of the feature extraction outputs reshaping  the resulting  tensor to $(W/4\times H/4\times C_\mathrm{7})$, (2) a self-attention operation which multiplies the  reshaped tensor by the output of a pooling, $1\times1$ convolution and sigmoid function which has the same concatenated tensor as its input and, (3) a $1\times1$ convolutional layer followed by a BatchNorm and LeakyRelu which acts as a bottleneck limiting the output to $C_6$ features. 

All implementation details, such as the number of features of each layer, are specified in Sect. \ref{sec:experimental_setup}. The experiments in Sect. \ref{sec:experiments} show how each part of this learning module contributes to improve 3D-MiniNet's performance.

 \subsubsection{2D Segmentation Module (MiniNet Backbone)}
Once the previous module has computed a $W/4\times H/4\times C_\mathrm{6}$ tensor, the 2D semantic segmentation is obtained running an efficient CNN (see MiniNet backbone in Fig. \ref{fig:mininet} for a visual description). 
Our module uses a FCNN instead of performing more MLP operations because convolutional layers have lower inference time when working on high dimensional spaces. 
Our FCNN is based on MiniNetV2 architecture \cite{alonso2020MininetV2}.  Our encoder performs $L_\mathrm{1}$ depthwise separable convolutions and $L_\mathrm{2}$ multi-dilation  depthwise separable convolutions. For the decoder, we use bilinear interpolations as upsampling layers. It performs $L_\mathrm{3}$ depthwise separable convolutions at  $W/4\times H/4$ resolution and $L_\mathrm{4}$  at  $W/2\times H/2$ resolution. Finally, a convolution is performed at  $W\times H$ resolution  to get the 2D semantic segmentation prediction. 

Similarly to MiniNetV2, we also include a second convolutional branch to extract fine-grained information, i.e., high-resolution low-level features. The input of this second branch is the spherical projection. 
The number of layers and features at each layer is specified in Sect. \ref{sec:settings}.

As a final step, the predicted 2D semantic segmentation labels are re-projected back into the 3D space ($\mathbb{R}^2 \xrightarrow{} \mathbb{R}^3$). For the points projected into the spherical representation, this reprojection is a straightforward step,
as it just implies assigning the semantic label predicted in the spherical projection. Nevertheless, the points that had not been projected into the spherical projection (one 2D coordinate can have more than one 3D point) have no semantic label. For these points,  the semantic label  of its corresponding 2D coordinate is assigned. As this issue may lead to miss-predictions, a post-processing method is performed to refine the results.

\subsection{Post-Processing}
In order to cope with the miss-predictions of non-projected 3D points, we follow Milioto et al. \cite{milioto2019rangenet++} post-processing method. 
All 3D points get a new semantic label  based on K Nearest Neighbors (KNN). 
The criteria for selecting the K nearest points is not based on the relative euclidean distances but on relative depth values. Besides, the search is narrowed down based on 2D spherical coordinate distances. Milioto et al. implementation is GPU-based and is able to run in 7ms keeping the frame-rate high.

\section{Experimental setup}
\label{sec:experimental_setup}

This section details the setup used in our experimental evaluation.

\subsection{Datasets}

\paragraph*{SemanticKITTI Benchmark}
The SemanticKITTI dataset \cite{behley2019semantickitti} is a recent large-scale dataset that provides dense point-wise annotations for the entire KITTI Odometry Benchmark \cite{geiger2012wekitti}.
The dataset consists of over 43000 scans from which over 21000 are available for training (sequences 00 to 10) and the rest (sequences 11 to 21) are used as test set.
The dataset distinguishes 22 different semantic classes from which 19 classes  are evaluated on the test set via  the official online platform of the benchmark. 
As this is the current most relevant and largest dataset of single-scan 3D LIDAR semantic segmentation, we perform our ablation study and our more thorough evaluation on this dataset.

\paragraph*{KITTI Benchmark}
SqueezeSeg \cite{wu2018squeezeseg} work provided semantic segmentation labels exported from the 3D object detection challenge of the KITTI dataset \cite{geiger2012wekitti}. It is a medium-size dataset split into 8057 training scans and 2791 validation scans.

\subsection{Settings}
\label{sec:settings}

\paragraph{3D Point Neighbor Search Parameters}
We set the resolution of the spherical projection to  $2048\times64$ for the SemanticKITTI dataset and $512\times64$ for the KITTI (same resolution than previous works to be able to make fair comparisons). 
We set a  $4\times4$ window size with a stride of $4$ and no zero-padding for our fast point neighbor search leading to 8192 groups of 3D points for the SemanticKITTI data and 2048 groups for the KITTI data.
Our projection module is fed with these groups and generates a learned representation of resolution  $512\times16$ for the SemanticKITTI configuration and $128\times16$ for the KITTI. 

\paragraph{Network Parameters}
\label{sec:network-parameters}
The full architecture and all its parameters are described in Fig. \ref{fig:mininet}. We considered three different configurations for evaluating the proposed approach: 3D-MiniNet, 3D-MiniNet-small, 3D-MiniNet-tiny.
 The number of features ($C_\mathrm{3}, C_\mathrm{4}, C_\mathrm{5}, C_\mathrm{6}$) for the projection module of the different 3D-MiniNet configurations are (24, 48, 96, 192) features for 3D-MiniNet,  (16, 32, 64, 128) for 3D-MiniNet-small and   (12, 24, 48, 96) for 3D-MiniNet-tiny.
The number of layers ($L_\mathrm{1}, L_\mathrm{2}, L_\mathrm{3}, L_\mathrm{4}$) of the FCNN backbone network are (50, 30, 4, 2) features for 3D-MiniNet,  (24, 20, 2, 1) for 3D-MiniNet-saml and   (14, 10, 2, 1) for 3D-MiniNet-tiny. $N_\mathrm{c}$ is the number of semantic classes of the dataset.

\paragraph{Post-processing Parameters}
For the K Nearest Neigbors post-process method \cite{milioto2019rangenet++}, we set as $7\times7$ the windows size of the neighbor search on the 2D segmentation and we set $K$ to 7.

\paragraph{Training protocol}
We train the different 3D-MiniNet configurations for 500 epochs with batch size of 3, 6 and  8 for 3D-MiniNet, 3D-MiniNet-small, and 3D-MiniNet-tiny respectively (different due to memory constraints). 
We use Stochastic Gradient Descent (SGD) optimizer with an initial learning rate of $4\cdot10^{-3}$ and a  decay of  0.99 every epoch.
For the optimization, we use the cross-entropy loss function, see eq. \ref{eq:lossf}.

\begin{equation}
\mathcal{L}= -\frac{1}{M}\sum_{m=1}^M\sum_{c=1}^C (\frac{f_{t}}{f_c})^{i} y_{c,m}\ln(\hat{y}_{c,m}),
\label{eq:lossf}
\end{equation}
\noindent where $M$ is the number of labeled points and $C$ is the number of classes. ${Y}_{c,m}$ is a binary indicator (0 or 1) of point $m$ belonging to a certain class $c$  and $\hat{y}_{c,m}$ is the CNN predicted probability of point $m$ belonging to a certain class $c$. This probability is calculated by applying the soft-max function to the networks' output. 
To account for class imbalance, we use the median frequency class balancing, as applied in SegNet~\cite{badrinarayanan2017segnet}. To smooth the resulting class weights, we propose to apply a power operation, $w_c =(\frac{f_{t}}{f_c})^{i}$, 
with $f_{c}$ being the frequency of class $c$ and $f_{t}$ the median of all frequencies. We set $i$ to 0.25.

\paragraph{Data augmentation}
During the training, we randomly rotate and shift the whole  3D point cloud. We randomly invert the sign for X and Z values for all the point cloud. We also drop some points.
The rotation angle is a Gaussian distribution with mean 0 and standard deviation  (std) of 40º.  The shifts we perform are Gaussian distributions with mean 0 and std of 0.35, 0.35 and 0.01 (meters) for the X, Y, Z axis (being Z the height).
The percentage of dropped points is a uniform distribution between 0 and 10.

\begin{table}[!t]
  \caption{Ablation study of the different parts of the projection module evaluated on the test set of SemantiKITTI.}
  \centering
   \resizebox{0.47\textwidth}{!}{
  \begin{tabular}{c|c c c c c c | c c c}
    \toprule[0.2em]
     &  Data &    & Local   & & Context   & Relative   & & & Params \\
    Method & Aug. &  Conv &  MLP  & Attention&  MLP  &  features  & mIoU & FPS &  (M) \\
    \toprule[0.2em]
     & &      \checkmark   &                 &    &    &      & 44.4& 73& 0.93  \\
     
     & \checkmark  &    \checkmark    &                 &    &    &      & 47.6& 73& 0.93  \\
  &  \checkmark   &           &   \checkmark       &     & & & 48.7 & 69& 0.93 \\
 3D-MiniNet & \checkmark    &         \checkmark      &   \checkmark           &      &  &  & 49.5&66 & 0.96 \\
 Small &\checkmark   &        \checkmark       &     \checkmark       &   \checkmark  &  &  &49.9 & 65&1.08  \\   
  & \checkmark   &       \checkmark           &      \checkmark          &    \checkmark     &   \checkmark    &  & 51.2& 61 &1.13  \\
    & \checkmark    &      \checkmark      &    \checkmark       &   \checkmark & \checkmark &  \checkmark & 51.8 &61 & 1.13  \\
    \bottomrule[0.1em]
  \end{tabular}
  } 
  \label{tab:ablation-study}
\end{table}

\begin{table*}[th]
\centering
\caption{
Results on single-scan test set in SemanticKITTI \cite{behley2019semantickitti}. Point-based methods: rows 1-4. 3D representations: row 5. Projection-based methods: rows 6-11.}
\label{tab:SemanticKITTI}
\resizebox{\textwidth}{!}{
\begin{tabular}{rccccccccccccccccccccccc}
\toprule[1.0pt]
Methods & Size & \rotatebox{90}{\textbf{mIoU}} & \rotatebox{90}{\textbf{Frame-rate (FPS)}}  & \rotatebox{90}{\textbf{Params(M)}}  & \rotatebox{90}{road IoU} & \rotatebox{90}{sidewalk IoU} & \rotatebox{90}{parking IoU} & \rotatebox{90}{other-ground IoU} & \rotatebox{90}{building IoU} & \rotatebox{90}{car IoU} & \rotatebox{90}{truck IoU} & \rotatebox{90}{bicycle IoU} & \rotatebox{90}{motorcycle IoU} & \rotatebox{90}{other-vehicle IoU} & \rotatebox{90}{vegetation IoU} & \rotatebox{90}{trunk IoU} & \rotatebox{90}{terrain IoU} & \rotatebox{90}{person IoU} & \rotatebox{90}{bicyclist IoU} & \rotatebox{90}{motorcyclist IoU} & \rotatebox{90}{fence IoU} & \rotatebox{90}{pole IoU} & \rotatebox{90}{traffic-sign IoU} \\
\toprule[1.0pt]
PointNet \cite{qi2017pointnet} & \multirow{5}{*}{50K pts} & 14.6 & 2 & 3 & 61.6 & 35.7 & 15.8 & 1.4 & 41.4 & 46.3 & 0.1 & 1.3 & 0.3 & 0.8 & 31.0 & 4.6 & 17.6 & 0.2 & 0.2 & 0.0 & 12.9 & 2.4 & 3.7 \\
SPG \cite{landrieu2018large} &  & 17.4 &0.2  & \textbf{0.25} & 45.0 & 28.5 & 0.6 & 0.6 & 64.3 & 49.3 & 0.1 & 0.2 & 0.2 & 0.8 & 48.9 & 27.2 & 24.6 & 0.3 & 2.7 & 0.1 & 20.8 & 15.9 & 0.8 \\
PointNet++ \cite{qi2017pointnet++} &  & 20.1  & 0.1 & 6 & 72.0 & 41.8 & 18.7 & 5.6 & 62.3 & 53.7 & 0.9 & 1.9 & 0.2 & 0.2 & 46.5 & 13.8 & 30.0 & 0.9 & 1.0 & 0.0 & 16.9 & 6.0 & 8.9 \\
RandLA-Net \cite{hu2020randla} &   & 53.9  & 22 & 1.24   &   90.7&73.7&60.3&20.4&86.9&\textbf{94.2}&\textbf{40.1}&26.0&25.8&\textbf{38.9}&81.4&\textbf{61.3}&\textbf{66.8}&\textbf{49.2}&\textbf{48.2}&7.2&56.3&\textbf{49.2}&47.7  \\
\toprule[0.5pt]

SPLATNet \cite{su2018splatnet} & \multirow{1}{*}{50K pts}   & 18.4 & 1 & 0.8 & 64.6 & 39.1 & 0.4 & 0.0 & 58.3 & 58.2 & 0.0 & 0.0 & 0.0 & 0.0 & 71.1 & 9.9 & 19.3 & 0.0 & 0.0 & 0.0 & 23.1 & 5.6 & 0.0  \\
\toprule[0.5pt]

SqueezeSeg \cite{wu2018squeezeseg} & \multirow{7}{*}{\begin{tabular}[c]{@{}c@{}}64x2048 px\end{tabular}} & 29.5 & 90  & 1 & 85.4 & 54.3 & 26.9 & 4.5 & 57.4 & 68.8 & 3.3 & 16.0 & 4.1 & 3.6 & 60.0 & 24.3 & 53.7 & 12.9 & 13.1 & 0.9 & 29.0 & 17.5 & 24.5 \\
DBLiDARNet \cite{dewan2019deeptemporalseg}&& 37.6 & --- & 2.8 & 85.8&59.3&8.7&1.0&78.6&81.5&6.6&29.4&19.6&6.5&77.1&46.0&58.1&23.7&20.1&2.4&39.6&32.6&39.1 \\

SqueezeSegV2 \cite{wu2019squeezesegv2} & & 39.7 & 83 & 1 & 88.6 & 67.6 & 45.8 & 17.7 & 73.7 & 81.8 & 13.4 & 18.5 & 17.9 & 14.0 & 71.8 & 35.8 & 60.2 & 20.1 & 25.1 & 3.9 & 41.1 & 20.2 & 36.3 \\

TangentConv \cite{tangentconv} &  & 40.9 & 0.3 & 0.4  & 83.9 & 63.9 & 33.4 & 15.4 & 83.4 & 90.8 & 15.2 & 2.7 & 16.5 & 12.1 & 79.5 & 49.3 & 58.1 & 23.0 & 28.4 & 8.1 & 49.0 & 35.8 & 28.5 \\
RangeNet21 \cite{milioto2019rangenet++} & & 47.4  &25 & 25 & 91.4 & 74.0 & 57.0 & 26.4 & 81.9 & 85.4 & 18.6 & 26.2 & 26.5 & 15.6 & 77.6 & 48.4 & 63.6 & 31.8 & 33.6 & 4.0 & 52.3 & 36.0 & 50.0 \\ 
RangeNet53 \cite{milioto2019rangenet++} & & 49.9 &13  & 50   &91.7 & 74.0 & 65.1 & 28.2 & 82.9 & 85.3 & 25.8 & 22.7 & 33.6 & 22.2 & 77.3 & 50.0 & 64.6 & 36.8 & 31.4 & 4.7 & 54.8 & 39.1 & 52.3 \\
RangeNet53-KNN \cite{milioto2019rangenet++} & & 52.2 &  12 & 50    & \textbf{91.8} & \textbf{75.2}&\textbf{65.0}&\textbf{27.8}&87.4&91.4&25.7&25.7&34.4&23.0&80.5&55.1&64.6&38.3&38.8&4.8&58.6&47.9&55.9  \\

\toprule[1.0pt]

\textbf{3D-MiniNet-tiny (Ours)} & & 46.9   &  \textbf{98} &  0.44  & 90.7 &70.7 &59.4 &20.0 &83.4 &82.0 &19.0 &29.3 &25.4 &20.8 &77.9 &50.6&  60.8 &35.1 &32.3 &3.2 &51.0 &32.7 &46.7  \\
\textbf{3D-MiniNet-small (Ours)} &\multirow{4}{*}{\begin{tabular}[c]{@{}c@{}}64x2048 px\end{tabular}}  &  51.8 & 61    & 1.13  & 91.5 & 72.3&  61.7 & 25.1 & 83.9  & 83.4 & 25.4 & 35.6  & 25.4&25.1  & 80.3 &  53.9 & 64.3  & 43.4 &  42.3 &20.7 & 53.0 &  36.4 & 50.3  \\ \vspace{1.2mm}
\textbf{3D-MiniNet (Ours)} & &  53.0 &     36 & 3.97  & 91.6 &74.0 &64.1 &25.9 &85.8 &85.2 &28.3 &37.9 &39.3 &28.8 &80.3 &54.5 &65.9 &43.8 &40.3 &14.4 &57.0 &37.9 &51.5  \\

\textbf{3D-MiniNet-tiny-KNN (Ours)} & &  49.0  & 55 &  0.44  & 90.7  &71.0 &59.5 &19.7 &86.4 &86.6 &19.2 &31.6 &27.8 &21.3 &80.0 &55.4 &61.4 &38.1 &35.0 &3.0 &53.7 &40.5 &51.0   \\ 
\textbf{3D-MiniNet-small-KNN (Ours)} & & 54.4&  40 & 1.13   &  91.5 & 72.7&  61.8 & 24.6 & 87.1  & 88.1  & 25.6& 39.3  & 38.0 & 25.6 &  82.5&  59.7 &  65.0 & 47.2 & 46.2  & \textbf{22.4}& 56.1 &  45.8 & 54.9  \\
\textbf{3D-MiniNet-KNN (Ours)} & &  \textbf{55.8}  & 28 & 3.97  &91.6   & 74.5 &64.2 &25.4 &\textbf{89.4} &90.5 & 28.5  &\textbf{42.3} &\textbf{42.1} &29.4&\textbf{82.8} & 60.8 & 66.7  & 47.8  & 44.1  &14.5 &\textbf{60.8 }&48.0 &\textbf{56.6} \\
\toprule[1.0pt]
 \multicolumn{15}{p{13cm}}{\footnotesize {Scans per second have been measured using a Nvidia gtx 2080ti}}\\ 
   \multicolumn{8}{p{10cm}}{\footnotesize {--- Not reported by the authors.}}
\end{tabular}
}
\end{table*}

\section{Results}
\label{sec:experiments}

\subsection{Ablation Study of the Projection Module}
The projection module is the main novelty from our approach. This subsection shows how each part helps to improve the learned representation.
For this experiment, we use 3D-MiniNet-small configuration.

Table \ref{tab:ablation-study} shows the ablation study of our proposed module, measuring the mIoU, speed and learning parameters needed with each configuration. The first row and baseline is working on the spherical projection using a convolution as the \textit{projection} method, i.e., just a downsampling in that case.

As the projection used is neither rotation nor shift invariant, performing this data augmentation helps to our network generalization as first row shows. Second row shows the performance using only $1\times N$ convolutions in the learning layers with the 5-channel input ($C_1$) used in RangeNet \cite{milioto2019rangenet++} which we establish as our baseline, i.e, our spatial feature extractor.
The third row shows the performance if we replace the $1\times N$ convolution for point-based operations, i.e, our local feature extractor. 
These results point that MLP operations work better for 3D points but take more execution time.
The fourth row combines both the convolution and local MLP operation. Combining  convolutions and MLP operations increases performance due to the different type of features learned by each type of operation as explained in Sect. \ref{sec:projection_module}.

The attention module also increases the performance with almost no extra computational effort. It reduces the feature space into a specified number of features, learning which features are more important.
The sixth row shows the results adding our context feature extractor. Context is also learned later through the FCNN via convolutions but here, the context feature extractor learns different context through with MLP operations. Context information is often very useful in semantic tasks, e.g., for distinguishing between a bicyclist, a cyclist and a motorcyclist. This context information gives a boost higher than the other feature extractors showing its relevance. 
Finally, increasing the number of features of each point with features relative to the point group ($C_2$) also leads  to better performance without decreasing the frame-rate and without adding any learning parameter.


\begin{table}[th]
\centering
\caption{Results on KITTI \cite{geiger2012wekitti} validation set.}
\label{tab:KITTI}
\resizebox{0.47\textwidth}{!}{
\begin{tabular}{rccccccc}
\toprule[1.0pt]
Methods & Size & \rotatebox{90}{\textbf{mIoU}} & \rotatebox{90}{\textbf{Frame-rate (fps)}}  & \rotatebox{90}{\textbf{Params(M)}}  & \rotatebox{90}{car IoU} & \rotatebox{90}{pedestrian IoU} & \rotatebox{90}{cyclist IoU}   \\
\toprule[1.0pt]
SqueezeSeg \cite{wu2018squeezeseg} & \multirow{5}{*}{\begin{tabular}[c]{@{}c@{}}64x512 px\end{tabular}}   & 37.2 & 227 & 1 & 64.6&21.8&25.1 \\
PointSeg \cite{wang2018pointseg}&   & 39.7 & 160 & --- &67.4&19.2 & 32.7\\ 
SqueezeSegv2 \cite{wu2019squeezesegv2} &    & 44.9 &143  &1 & 73.2 &27.8 &33.6  \\
LuNet \cite{lunet}&   & 55.4 & 67* & 23.4 &72.7 &46.9 &46.5 \\ 
DBLiDARNet \cite{dewan2019deeptemporalseg}&   & 56.0 & --- & 2.8 &75.1 &47.4 &45.4 \\ 
\toprule[1.0pt]

\textbf{3D-MiniNet-tiny (Ours)}&   & 45.5 &  \textbf{245} & \textbf{0.44} &  69.6 & 37.5 & 29.5  \\
\textbf{3D-MiniNet-small (Ours)} &\multirow{1}{*}{\begin{tabular}[c]{@{}c@{}}64x512 px\end{tabular}}   & 50.6 & 161 & 1.13  & 74.4 & 40.7 & 36.7 \\
\textbf{3D-MiniNet (Ours)} &   & \textbf{58.0} & 92 & 3.97  & \textbf{75.5 }&\textbf{49.6}  &\textbf{48.9} \\
 
\toprule[1.0pt]
 \multicolumn{8}{p{10cm}}{\footnotesize {Scans per second have been measured using a Nvidia gtx 2080ti}}\\
  \multicolumn{8}{p{10cm}}{\footnotesize {* Offline neighboring point search is not taken into account.}}\\
  \multicolumn{8}{p{10cm}}{\footnotesize {--- Not reported by the authors.}}
  
\end{tabular}
}
\end{table}

\subsection{Benchmarks results}
This subsection presents  quantitative and qualitative results of  3D-MiniNet and comparisons with other relevant works.

\begin{figure*}[!tb]
\centering
\includegraphics[width=1\linewidth]{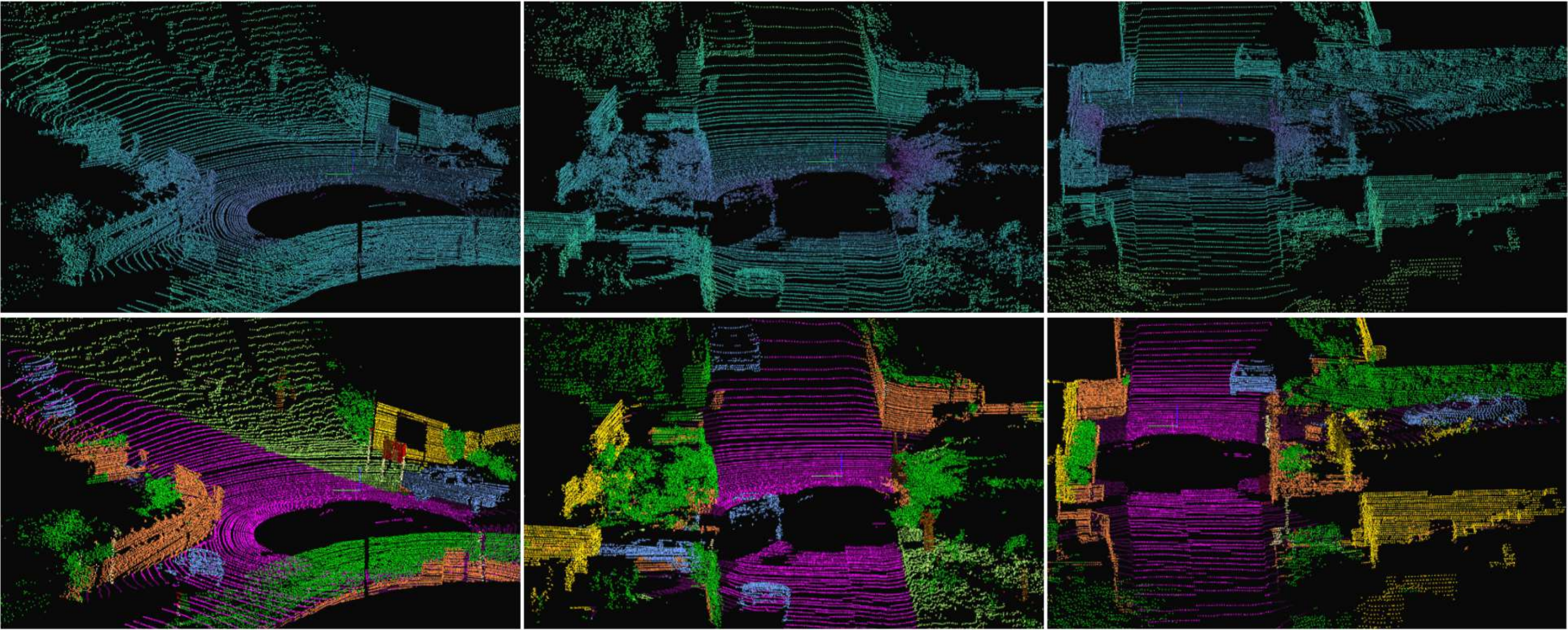}
\caption{3D-MiniNet LIDAR semantic segmentation predictions on the SemanticKITTI   benchmark (test sequence 11). LIDAR point cloud are on top  where color represents depth. Predictions are on bottom where color represents semantic classes: cars in blue, road in purple, vegetation in green, fence in orange, building in yellow and traffic sign in red. \href{https://www.youtube.com/watch?v=5ozNkgFQmSM}{For the full video sequence, go to https://www.youtube.com/watch?v=5ozNkgFQmSM}.
Best viewed in color. }
\label{fig:results}
\end{figure*} 

\paragraph{Quantitative Analysis} 
Table \ref{tab:SemanticKITTI} compares our method with several point-based approaches (rows 1-4), 3D representation methods (row 5) and projection-based approaches (rows 6-11) measuring the mIoU, the processing speed (FPS) and the number of parameters required by each method. 
As we can see, point-based methods for semantic segmentation of LIDAR scans tend to be slower than projection ones without providing better performance. As LIDAR sensors such as Velodyne usually work at 5-20 FPS, only RandLA-Net and projection-based approaches are currently able to process in real time the full amount of data made available by the sensor.

Looking at the different configurations of 3D-MiniNet, it gets state-of-the-art using fewer parameters and being faster (3D-MiniNet-small-KNN) beating both RandLANet (point-based method), SPLATNet (3D representation) and RangeNet53-KNN (projection-based). Besides, 3D-MiniNet-KNN configuration is able to get even better performance although it needs more parameters than RandLANet.
If efficiency can be traded off for performance, smaller versions of Mininet also obtain better performance metrics at higher frame-rates. 3D-MiniNet-tiny is able to run at 98 fps and, with only a $9\%$ drop in mIoU ($46.9\% $ compared to the $29\%$ of SqueezeSeg version that runs at 90 fps).

The post-processing method applied \cite{milioto2019rangenet++} shows its effectiveness improving the results the same way it improved RangeNet. 
This step is crucial to correctly process points that were not included in the spherical projection, as discussed in more detail in Sect. \ref{sec:method}.

The scans of the KITTI dataset \cite{geiger2012wekitti} have a lower resolution (64x512) as we can see in the evaluation reported in Table \ref{tab:KITTI}.  
3D-MiniNet also gets state-of-the-art performance on LIDAR semantic segmentation on this dataset.
Our approach gets considerably better performance than SqueezeSeg versions (+10-20 mIoU). 
3D-MiniNet also gets better performance than LuNet and  DBLiDARNet which were the previous best methods on this dataset.

Note that in this case, we did not evaluate the KNN post-processing since this dataset only provides 2D labels.  

The experiments show that projection-based methods are more suitable for the LIDAR semantic segmentation with a good speed-performance trade-off. Besides, better results are obtained when including point-based operations to extract both context and local information from the 3D raw points into the 2D projection.

\paragraph{Qualitative Analysis} Fig. \ref{fig:results} shows a few examples of 3D-MiniNet inference on test data. 
The supplementary video includes inference results on a full sequence\footnote{\url{https://www.youtube.com/watch?v=5ozNkgFQmSM}}.
As test ground-truth is not provided for the test set (evaluation is performed externally on the online platform), we can only show visual results with no label comparison.  

Note the high quality results on our method in relevant classes such as cars, as well as in challenging classes such as traffic signs.
In the \href{https://www.youtube.com/watch?v=5ozNkgFQmSM}{supplementary video} we can also appreciate some of the 3D-MiniNet failure cases. As it could be expected, the biggest difficulties happen  distinguishing between classes with similar geometric shapes and structures like building and fences.

\section{CONCLUSIONS}

In this work, we propose 3D-MiniNet, a fast and efficient approach for 3D LIDAR semantic segmentation. 
3D-MiniNet projects the 3D point cloud into a 2-Dimensional space and then learns the semantic segmentation using a fully convolutional neural network.
Differently from common projection-based approaches that perform a predefined projection, 3D-MiniNet learns this projection from the raw 3D points, learning both local and context information from point-based operations, showing very promising and effective results. 
Our ablation study shows how each part of the proposed approach contributes to the learning of the representation. 
We validate our approach on the SemanticKITTI and KITTI public benchmarks. 3D-MiniNet gets state-of-the-art results while being faster and more efficient than previous methods.

{
\bibliographystyle{IEEEtran}
\bibliography{biblio}
}

\end{document}